\documentclass[11pt]{article}
\usepackage{booktabs}
\usepackage[table]{xcolor}
\usepackage{multirow} 
\usepackage{colortbl} 
\usepackage{caption} 
\usepackage[final]{acl}
\usepackage{times}
\usepackage{url}
\usepackage{amsmath} 
\usepackage{cleveref}
\usepackage{latexsym}
\usepackage{graphicx}
\usepackage{array}
\usepackage{tabularx}
\usepackage{enumitem}
\usepackage{adjustbox}
\usepackage{booktabs}
\usepackage{caption}
\usepackage{geometry}
\newcommand{\stitle}[1]{\vspace{1ex} \noindent{\bf #1.}}
\usepackage{xspace}
\usepackage{yfonts}

\usepackage{algorithm}
\usepackage{algorithmic}
\usepackage{microtype}
\crefformat{section}{\S#2#1#3}
\crefformat{subsection}{\S#2#1#3}
\crefformat{subsubsection}{\S#2#1#3}
\crefrangeformat{section}{\S\S#3#1#4 to~#5#2#6}
\crefmultiformat{section}{\S\S#2#1#3}{ and~#2#1#3}{, #2#1#3}{ and~#2#1#3}

\Crefformat{figure}{#2Fig.~#1#3}
\Crefmultiformat{figure}{Figs.~#2#1#3}{ and~#2#1#3}{, #2#1#3}{ and~#2#1#3}
\Crefformat{table}{#2Tab.~#1#3}
\Crefmultiformat{table}{Tabs.~#2#1#3}{ and~#2#1#3}{, #2#1#3}{ and~#2#1#3}
\Crefformat{appendix}{#2Appx.~\S#1#3}
\crefformat{algorithm}{Alg.~#2#1#3}

\newcommand{\modelname}{\texttt{\textbf{R2D2}}{}\xspace}

\newcommand{\modelmeaning}{\texttt{\textbf{R}}emembering, \texttt{\textbf{R}}eplaying, and \texttt{\textbf{D}}ynamic \texttt{\textbf{D}}ecision
Making\xspace}

\title{\includegraphics[scale=0.35]{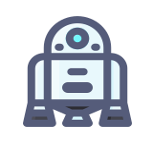} \modelname: Remembering, Replaying and Dynamic Decision Making\\ with a Reflective Agentic Memory}

\author{
Tenghao Huang\textsuperscript{1\thanks{Work done during Tenghao's internship at IBM research.}} \quad
Kinjal Basu\textsuperscript{2} \quad 
Ibrahim Abdelaziz\textsuperscript{2} \quad \\
\textbf{Pavan Kapanipathi\textsuperscript{2} \quad 
Jonathan May\textsuperscript{1} \quad
Muhao Chen\textsuperscript{3}} \\
\textsuperscript{1}University of Southern California,
\textsuperscript{2}IBM Research,
\textsuperscript{3}University of California, Davis \\
\texttt{tenghaoh@usc.edu}
}

\begin{document}
\maketitle
\begin{abstract}
The proliferation of web agents necessitates advanced navigation and interaction strategies within complex web environments. Current models often struggle with efficient navigation and action execution due to limited visibility and understanding of web structures. Our proposed \modelname framework addresses these challenges by integrating two paradigms: Remember and Reflect. The Remember paradigm uses a replay buffer that aids agents in reconstructing the web environment dynamically, thus enabling the formulation of a detailed ``map'' of previously visited pages. This helps in reducing navigational errors and optimizing the decision-making process during web interactions. Conversely, the Reflect paradigm allows agents to learn from past mistakes by providing a mechanism for error analysis and strategy refinement, enhancing overall task performance. We evaluate \modelname using the WebArena benchmark, demonstrating substantial improvements over existing methods, including a 50\% reduction in navigation errors and a threefold increase in task completion rates. Our findings suggest that a combination of memory-enhanced navigation and reflective learning promisingly advances the capabilities of web agents, potentially benefiting various applications such as automated customer service and personal digital assistants.
\end{abstract}

\section{Introduction}

Web agents---autonomous AI agents designed to navigate and perform natural language-described tasks within web environments---have become increasingly integral to applications such as online customer service \cite{huang-etal-2025-crmarena}, automated data retrieval \cite{huang-etal-2024-autoscraper}, and personalized digital assistants.\footnote{\url{https://www.anthropic.com/news/3-5-models-and-computer-use}} These agents interact with complex web interfaces to execute user-described tasks, often emulating human actions like clicking buttons, filling forms, and extracting information \cite{pmlr-v70-shi17a, liu2018reinforcement, c3260a6257764eadb1fc9439b8c8ae9b, zhouwebarena}. 
\begin{figure}[t!]
    \centering
\small    \includegraphics[width=0.75\linewidth]{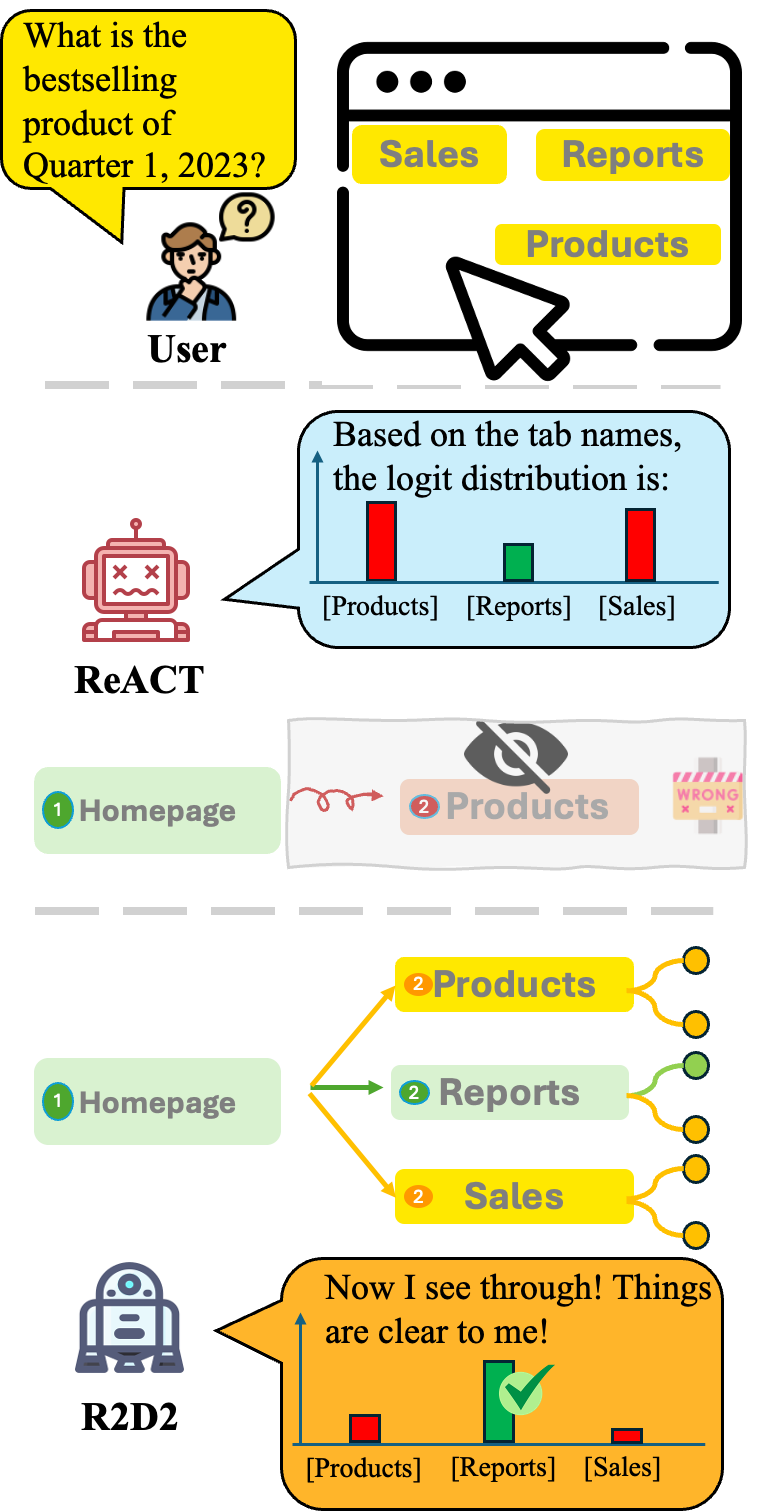}
    
    \caption{Traditional methodologies conceptualize web navigation within the framework of an Unknown MDP. The ReACT agent operates under high uncertainty due to incomplete information regarding the outcomes of its actions, leading to erroneous navigational paths and impeding effective task resolution. \modelname transforms the task into a Known MDP, improving robustness. }
    \vspace{-5mm}
    \label{fig:intro_figure}
\end{figure}

\begin{figure*}[t!]
    \centering
    \small
    \includegraphics[width=0.8\linewidth]{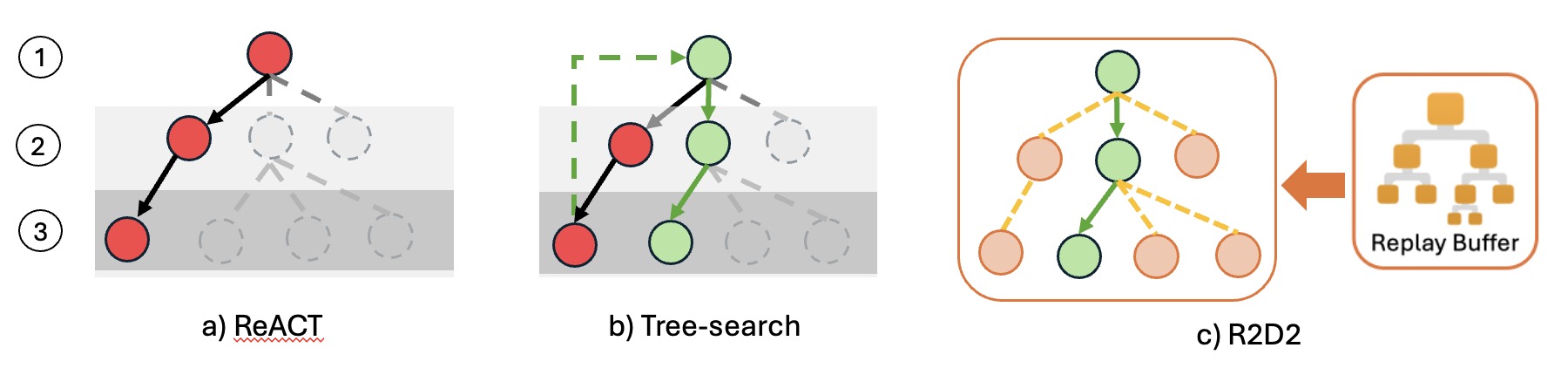}
    \vspace{-3mm}
    \caption{This diagram represents various approaches for web agents framed as a search problem, where each node symbolizes a webpage. (a) ReACT: The agent chooses the best immediate actions without any proactive strategy. (b) Tree-search with reflections: the agent investigates different routes by actively navigating websites and allows for reversing direction (shown by dashed arrows). 
    Both a) and b) approaches are Unknown-MDP-based. At each timestep, these agents' observation space is constrained, which typically results in suboptimal or inefficient results. c) \modelname: Our proposed framework constructs the search space, leveraging stored state information from a replay buffer. By transforming the task into a Known MDP, \modelname enhances its ability to navigate and interact with web interfaces.}
    \vspace{-3mm}
    \label{fig:conceptual_diff}
\end{figure*}

Despite recent advancements in web agents' capabilities, a persistent challenge remains: agents frequently fail to navigate effectively within intricate web environments as illustrated in \Cref{fig:intro_figure}.

The fundamental challenges associated with previous methodologies are twofold. First, these approaches model web navigation as an Unknown Markov Decision Process (MDP), wherein the agent has limited visibility into the consequences of its actions, often leading to suboptimal performance outcomes. Second, prior methods engage in complex reasoning during the inference phase while observing a stream of experiences, and in their simplest forms, they discard incoming trajectories immediately after a single episode. This rapid forgetting of potentially valuable experiences impedes the agent's capacity to leverage useful information for future decision-making.

Meanwhile, a primary obstacle in enhancing web agent performance lies in navigation-related failures, which account for approximately 60\% of their operational errors (as illustrated in \Cref{fig: error_analy}). These failures occur when agents become disoriented within the web environment, preventing them from reaching the target webpages necessary to execute desired tasks. Such navigation inefficiencies significantly hinder the overall effectiveness of web agents. The remaining 40\% of errors stem from execution failures and edge cases, where agents either misinterpret user intentions or mishandle specific web elements. 

Inspired by cognitive studies showing that humans excel at complex tasks by iteratively refining strategies based on feedback \cite{palenciano2021exploring, zenkri2024extracting}, as well as by approaches in robotics for structured exploration of unfamiliar spaces \cite{thrun2002probabilistic}, we propose the \modelname (\modelmeaning) framework that enhances both navigation and task execution for web agents. 
Our method transforms the task from an Unknown-MDP into a Known MDP by introducing the Remember Paradigm. It leverages a structured replay buffer of the agent's experiences that guides the agent to more promising avenues \cite{blundell2016model, schaul2016prioritizedexperiencereplay, parisotto2017neuralmapstructuredmemory, savinov2018semiparametrictopologicalmemorynavigation}. At a high level, our approach enables the agent to record and recall previously visited pages—essentially constructing a dynamic map of the web environment—and then leverage this knowledge to refine its strategies. By converting the agent's experience into a well-organized search space, we empower it to identify reliable navigation routes to target resources rather than re-deriving them from scratch during inference. This reduces the computational overhead at test time and helps avoid repetitive or unproductive exploration.

To enable continual improvement based on both successes and failures, \modelname incorporates the Reflect Paradigm. Previous efforts in this domain often focus narrowly on immediate, execution-level errors and struggle with more pervasive navigation challenges \cite{shinn2023reflexionlanguageagentsverbal, panautonomous, wang-etal-2024-devils}. In contrast, our method leverages the refined, known search space described earlier to minimize navigational missteps, allowing the reflection mechanism to operate more effectively on remaining execution problems. By reducing the burden of basic wayfinding, the agent's reflective capabilities can more efficiently identify and correct subtle issues, ultimately leading to a higher overall success rate on complex web tasks.

Our proposed method diverges from traditional techniques by providing a more comprehensive and structured representation of the agent's historical experiences. Instead of simply recalling past states \cite{agashe2024agentsopenagentic} or relying on on-the-fly reasoning \cite{koh2024tree, zhou2024language}, our approach organizes the agent's accumulated experiences into a coherent and reusable resource that effectively guides future decisions. 
We evaluate our approach using the \textsc{WebArena} benchmark \cite{zhouwebarena}, where it achieves substantial gains compared to baseline models, including approximately a 50\% reduction in navigation errors and a threefold increase in overall task completion rates. Moreover, \modelname outperforms state-of-the-art methods by 17\%, thereby demonstrating a more robust and informed capability for executing complex web-based tasks.

\section{Related Works}

\stitle{Enhancing Web Agents} Existing research has shown that language models, without intervention, struggle to express linguistic intent in formal instruction that can control an extra-linguistic environment, such as web site navigation \cite{pmlr-v70-shi17a, liu2018reinforcement, c3260a6257764eadb1fc9439b8c8ae9b, zhouwebarena, deng2023mind2web}. This inadequacy stems primarily from the intrinsic challenges associated with perception, strategic planning, and task execution in the intricate web environments.

To mitigate these challenges, enhancements to web agents have been characterized as one of three principal strategies: 
(1) \textit{Perception Alignment}: This strategy aims to augment agents' capabilities in interpreting graphical user interface elements by integrating multimodal data from webpages, enhancing both textual and visual comprehension \cite{gou2025uground, liuharnessing}. 
(2) \textit{Post-hoc Reflection}: Studies indicate that enabling agents to engage in reflective practices post-interaction can facilitate learning from historical trajectories, thereby improving future task executions \cite{shinn2023reflexionlanguageagentsverbal, song-etal-2024-trial, panautonomous, wang-etal-2024-devils}. (3) \textit{Online Search Algorithms}: This involves the adoption of sophisticated search algorithms, including Monte Carlo Tree Search and other tree-based exploration methods, integrated with high-level planning driven by LLM-derived knowledge \cite{koh2024tree, meng-etal-2024-llm, zhang2025webpilot}. Furthermore, \citealt{gu2024llmsecretlyworldmodel} discusses speculative planning that leverages simulations of world models.

Despite these enhancements, the performance of current web agents are constrained by the assumptions of Unknown MDP, where the potential outcomes of actions are not available. In contrast, this paper proposes a novel approach where we reconstruct the web environment's structure based on an agent's exploratory actions, thereby furnishing it with outcome information crucial for making informed and grounded decisions.

\stitle{Continual Exploration of the Agentic Environment} 
Tasking agents to explore an unknown environment has been an active research direction  \cite{brohan2023rt1roboticstransformerrealworld}. Recent studies have focused on how agents abstract experiences into actionable skills, a development that is becoming increasingly central to advancements in this field \cite{Wang2023VoyagerAO, wang2023jarvis1openworldmultitaskagents,liu2024autodan}. 
Within the domain of web-based agents, these skills are often conceptualized as workflows. \citet{sodhi2024step} have introduced a novel framework that leverages human-engineered workflows to compose policies to tackle web tasks. Although this improves agent performance, manually crafting such  workflows can be a tedious process. 

Unlike previous strategies that rely solely on high-quality successful trajectories or hand-crafted workflows, \modelname introduces a two-part mechanism that continuously learns from the full range of agent experiences, including failed attempts. 
\modelname moves beyond the limitations of purely Unknown-MDP-based assumptions and handcrafted workflows, resulting in more informed, robust decision-making and improved overall performance.

\begin{figure*}[t!]
    \centering
    \includegraphics[width=0.8\linewidth]{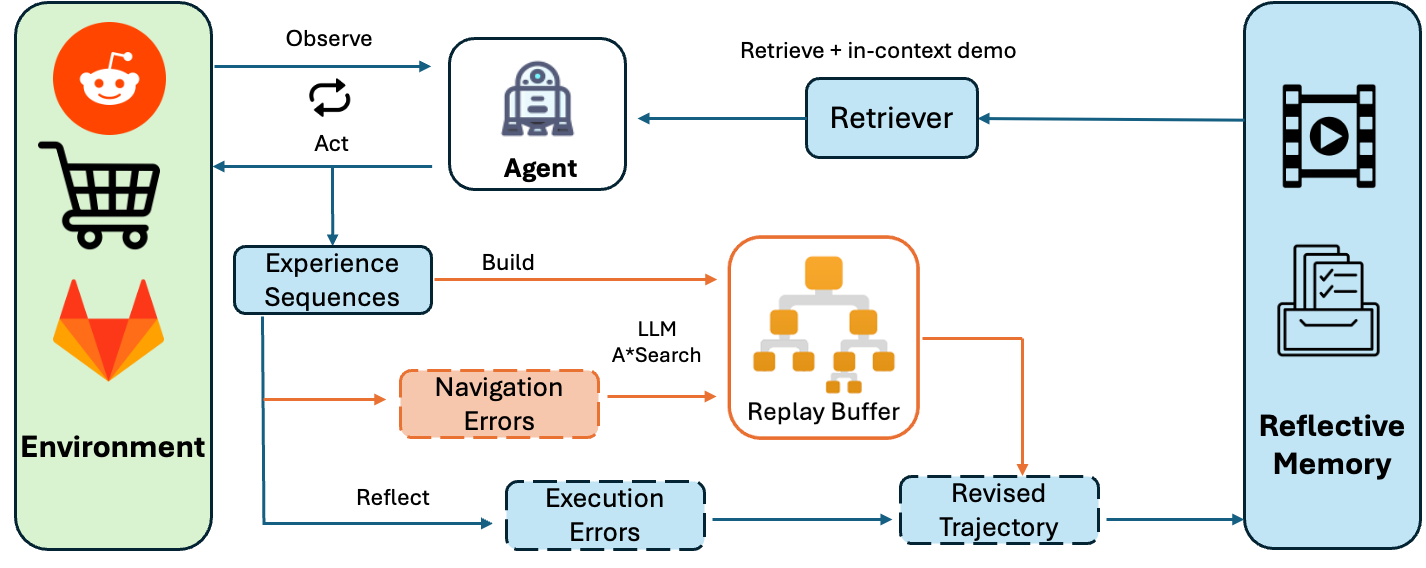}
    
    \caption{An overview of the \modelname architecture, highlighting the Remember and Reflect Paradigms. The Remember paradigm constructs a structured replay buffer from previous observations, enabling the agent to use past episodic data through A* search for navigation. Meanwhile, the Reflect paradigm diagnoses errors and generates corrective insights, which are then stored in a reflective memory for future decision-making processes.}

    \label{fig:pipeline}
\end{figure*}
\section{Method}


In this section, we present our framework, which tackles complex web navigation tasks by integrating two paradigms: Remember and Reflect.
The Remember paradigm constructs a structured replay buffer from past observations (\Cref{ssec:remember}), while the Reflect paradigm diagnoses and corrects errors in failed trajectories (\Cref{ssec:reflect}). We then introduce mechanisms of the reflective memory (\Cref{ssec:reflective_memory}). Finally, we illustrate how these paradigms interact to improve the agent's performance through retrieval and in-context learning demonstrations (\Cref{ssec:inference}).

\subsection{Method Overview}\label{ssec:overview}

Given a user's task query \( q \) and an initial observation \( o_0 \) from the environment, the agent must produce a sequence of actions to address \( q \). We define an \emph{episode} as the process where the agent starts from \( o_0 \) and executes a trajectory \( t \). Let \( t = \{ a_1, a_2, \ldots, a_H \} \) be the trajectory of length \( H \), where each \( a_h \) is an action at step \( h \). After each action \( a_h \), the agent receives an observation \( o_h \), thereby forming an observation sequence \( O = \{ o_1, o_2, \ldots, o_H \} \). Consider \( N \) distinct user queries \(\{ q_1, q_2, \ldots, q_N \}\), each associated with its own episode and observation sequence \( O_i = \{ o_{i,1}, o_{i,2}, \ldots, o_{i,H} \} \). Across all \( N \) user queries, \modelname forms the union of these observations:$ \mathcal{O}_{\text{all}} = \bigcup_{i=1}^N O_i.$

\modelname addresses errors in trajectories by categorizing them into two distinct types: (1) \stitle{Navigation failure}  We define \( O^*\) as the \emph{key observations} essential for successfully addressing the query \( q \). After performing the trajectory $t$, the observation sequence may not contain all the key observations, which leads to agents' navigation failure in the environment, formally $O \cap O^* \neq O^*$. The agent fails because of incomplete information or tools to address the user query. (2) \stitle{Execution failure} After performing $t$, $O \cap O^* = O^*$. In other words, navigation was successful, since the proper path was followed, but the agent still failed to address the user query.

The \emph{Remember} paradigm aims to build a replay buffer from $\mathcal{O}_{all}$, allowing the agent to store and revisit past observations and experiences \cite{blundell2016model, schaul2016prioritizedexperiencereplay, parisotto2017neuralmapstructuredmemory, savinov2018semiparametrictopologicalmemorynavigation}. The \emph{Reflection} paradigm then corrects trajectories that failed due to execution issues, providing explicit rationales for these execution failures. The successful and corrected trajectories and their rationales are stored in a \emph{reflective memory} for the agent's future reference. Finally, the \emph{Retriever} leverages this reflective memory by selecting relevant corrected trajectories as in-context demonstrations, thereby continually improving the agent's performance.

\subsection{Remember Paradigm}
\label{ssec:remember}
\modelname first determines\footnote{Please refer to \Cref{fig:error_classification_p} in Appendices for the detailed prompt.} the error type of a failed trajectory $t$. 
For all navigation failures, \modelname performs search within the replay buffer to correct their navigation behaviors. 

\stitle{Building the Replay Buffer}
To effectively represent the web environment and enable its systematic reconstruction from the observation sequence $\mathcal{O}_{\text{all}}$, \modelname structures the replay buffer as a directed graph \(G = (O, E)\). Here, the vertex set \(O\) includes the root node \(o_0\), which corresponds to the homepage observation, and each subsequent vertex represents another webpage observation \(o_i\). The edge set \(E\) consists of tuples \(((o_i, o_j),a)\) where each edge corresponds to an action \(a\) that transitions the agent from one observation \(o_i\) to another observation \(o_j\). Due to the noisy and dynamic nature of web pages, \modelname stores the differences between consecutive observations at each vertex rather than the full webpage state.

\stitle{{$\mathbf{A}^*$}{} Search within the Buffer} 
\label{ssec:astar}
Our search algorithm employs a best-first search ($A^*$) strategy \cite{Hart1968, meng-etal-2024-llm} to navigate and evaluate web environments effectively. Instead of expanding nodes level-by-level, \modelname incorporates a heuristic that guides the search toward potentially relevant and promising webpages more efficiently \cite{bonet1999planning, guez2018learning, moldovan2012safeexplorationmarkovdecision}. This heuristic, provided by the LLM, estimates the relevance and utility of exploring a particular webpage node, thereby reducing unnecessary expansions and focusing on paths that are more likely to yield correct information.

In $A^*$ search, each node \( o \) in the replay buffer graph is associated with both a cost (e.g., the depth from the start node or the number of steps taken) and a heuristic \( h(o) \), which estimates how close \( o \) is to a relevant webpage that can answer the query \( q \). We derive the heuristic by prompting the LLM to assess the likelihood that the subtree rooted at \( o \) will yield information relevant to \( q \). $A^*$ search proceeds by maintaining a priority queue that selects which node to expand next based on the sum of the cost-to-come and the heuristic estimate. Webpages that are deemed relevant are added to a candidate queue, prioritizing content that potentially answers the query, while non-relevant pages are bypassed to streamline the search. This exploration continues until all reachable nodes have been evaluated. Subsequently, for each candidate node in the queue, paths are constructed back to the root, mapping feasible routes that could satisfy the query. The LLM then ranks these paths based on relevance and utility, culminating in the selection of the optimal trajectory \( P^* \). 

\begin{algorithm}
\small
\caption{Optimized Web Search Using $A^*$ and Language Model}
\label{alg:websearch}
\label{alg:websearch_heuristic}
\begin{algorithmic}[1]
\REQUIRE User query \( q \), Initial observation \( o_0 \)
\ENSURE Optimal solution trajectory \( P^* \) for  \( q \)

\STATE Initialize replay buffer graph \( G = (O, E) \)
\STATE Add root node \( o_0 \) to \( O \) 

\STATE Initialize priority queue \( Q_{A^*} \) with \( (o_0, f(o_0) = h(o_0)) \) 
\STATE Initialize candidate queue \( Q_{cand} \leftarrow \emptyset \)

\WHILE{\( Q_{A^*} \) is not empty}
    \STATE \( (o_i, f(o_i)) \leftarrow \text{dequeue}(Q) \)
    
    \IF{\( \text{IsRelevant}(o_i, q, LLM) \)}
        \STATE \( Q_{cand} \leftarrow \text{enqueue}(Q_{cand}, (o_i, f(o_i))) \)
    \ENDIF
    
    \FOR{each action \( a \) available at \( o_i \)}
        \STATE \( o_j \leftarrow \text{Transition}(o_i, a) \) \COMMENT{Obtain next observation via action \( a \)}
        
        \IF{\( o_j \) not in \( \text{Visited} \)}
            \STATE \( h(o_j) \leftarrow \text{Heuristic}(o_j, q) \)
            \STATE \( f(o_j) \leftarrow f(o_i) + h(o_j)  \)
            \STATE \( Q_{A^*}  \leftarrow \text{enqueue}(Q, (o_j, f(o_j))) \)

        \ENDIF
    \ENDFOR
\ENDWHILE

\STATE Initialize trajectory set \( \mathcal{T} \leftarrow \emptyset \)

\FOR{each \( o_i \) in \( Q_{cand} \)}
    \STATE \( t \leftarrow \text{Backtrack}(o_i) \) \COMMENT{Generate trajectory from \( o_0 \) to \( o_i \) by following parent pointers}
    \STATE \( \mathcal{T} \leftarrow \mathcal{T} \cup \{t\} \)
\ENDFOR

\STATE \( P^* \leftarrow \text{RankAndSelectOptimal}(\mathcal{T}, q) \) \COMMENT{Use LLM to rank trajectories based on relevance and utility}

\RETURN \( P^* \)

\end{algorithmic}
\end{algorithm}

This $A^*$-base approach enables more informed and targeted exploration of the replay buffer.
By guiding the agent through the environment with a heuristic informed by the LLM, \modelname narrows down the search space and accelerates the discovery of relevant information. This ultimately results in faster and more accurate trajectory generation, effectively addressing complex user queries.

\subsection{Reflect Paradigm}
\label{ssec:reflect}

We discussed details of how we address navigation failures in \Cref{ssec:remember}. We now discuss using reflection techniques to address execution errors. 

The reflection process is designed to enhance the system's capability to learn from mistakes within trajectories rather than only successes \cite{madaan2023selfrefine, shinn2023reflexionlanguageagentsverbal}. 
When the failure reason of a trajectory $t$ is classified as an execution failure, we prompt the LLM to identify the first erroneous action $a_i$. 
The trajectory is then truncated to include only the actions before the error point, $\{a_1, a_2, \ldots, a_{i-1}\}$, which are considered correct. 
Following this, a detailed reflection on the erroneous action $a_i$ is generated, providing a rationale for its failure and potential strategies for avoidance in the future. 
This reflection, along with the truncated trajectory, is stored in the reflective memory \cite{weston2015memorynetworks, mirowski2017learningnavigatecomplexenvironments, wayne2018unsupervised} that is to be introduced in \Cref{ssec:reflective_memory}. 


\subsection{Reflective Memory Mechanism}
\label{ssec:reflective_memory}
We introduce the reflective memory mechanism that stores corrected and truncated trajectories for future retrieval. The reflective memory is structured as a key-value store:

\stitle{Key-Value Architecture} The reflective memory mechanism functions as a key-value store where each user query is encoded into a unique query vector serving as the key, encapsulating the query's semantic intent for efficient retrieval via vector similarity metrics. The corresponding value comprises a truncated and corrected trajectory, as described in \Cref{ssec:overview}, along with reflective insights. Specifically, for execution failures, only steps up to the first error are stored, while for navigation failures, corrected trajectory segments are retained once identified. During inference, a new query vector is generated and matched against existing keys to retrieve the most relevant trajectories.




\stitle{Basic Operations} In alignment with conventional memory module architectures, the reflective memory mechanism defines two basic operations: (1) \textit{Lookup.} Given a query, the memory retrieves the value(s) associated with the closest key vectors.
(2) \textit{Update.} If a newly truncated trajectory provides a more accurate or enriched reflection for an existing query, the memory updates the current value. \modelname uses an LLM to make such decision. Please refer to \Cref{fig:update_eval_p} in the Appendix for an example.


\subsection{Paradigm Coordination and Inference}\label{ssec:inference}

\stitle{Exploration Phase}
Using a ReACT agent \cite{yao2023reactsynergizingreasoningacting}, \modelname processes user queries and collects observational data to build \( \mathcal{O}_{\text{all}} \). Trajectories are classified and corrected via Remember and Reflect paradigms, then stored in the memory.

\stitle{Inference Phase}
During inference, user queries are encoded into vectors and matched against reflective memory to retrieve relevant trajectories as in-context demonstrations \cite{karpukhin-etal-2020-dense, brown2020languagemodelsfewshotlearners}. These demonstrations guide the agent's response. Failed trajectories undergo reflection, and the memory is updated to improve future performance.
This coordination allows \modelname to leverage past experiences and reflections, ensuring continuous learning and enhanced handling of complex queries.

\section{Experiments}
In this section, we evaluate the proposed \modelname framework for web agent tasks and compare it with baseline methods. We first delve into the details of our experimental setup (\cref{ssec: setup}), discuss the results obtained (\cref{ssec: results}), and perform ablation studies to understand the strengths of different components (\cref{ssec: ablation}). Furthermore, we provide a comprehensive error analysis (\Cref{ssec: error_analy_4}).

\subsection{Experimental Setup}
\label{ssec: setup}

\begin{table*}[ht]
\small
\centering
\setlength\tabcolsep{9pt} 
\begin{tabular}{@{} l ccccc c@{}}
\toprule
\textbf{Method} & \multicolumn{5}{c}{\textbf{Tasks}} & \textbf{Total SR} \\
\cmidrule(lr){2-6}
& \textbf{CMS} & \textbf{Reddit} & \textbf{Shopping} & \textbf{Map} & \textbf{GitLab} & \\
\midrule

ReACT\cite{yao2023reactsynergizingreasoningacting}\textsuperscript{\dag}  & -- & -- & -- & -- & -- & 13.1\% \\
Tree-Search \cite{koh2024tree}\textsuperscript{\dag} & 17\% & 11\% & 28\% & 26\% & 13\% & 19.0\% \\

AutoEval \cite{panautonomous}\textsuperscript{\ddag} & -- & -- & -- & -- & -- & 20.2\% \\
LATS \cite{zhou2024language}\textsuperscript{\ddag} & 15\% & \textbf{25\%} & 30\% & 27\% & 17\% & 22.5\%\\
AR \cite{wang-etal-2024-devils}\textsuperscript{\ddag} & 16\% & 24\% & 32\% & 27\% & 18\% & 23.4\% \\
BrowserGym \cite{drouin2024workarenacapablewebagents}\textsuperscript{\ddag} & -- & -- & -- & -- & -- & 23.5\% \\
\textbf{R2D2}\textsuperscript{\dag} & \textbf{30\%} & 21\% & \textbf{36\%} & \textbf{28\%} & \textbf{28\%} & \textbf{27.3\%} \\
\bottomrule
\end{tabular}
\caption{
Performance comparison across multiple web-based tasks. Reported success rates (SR) are organized by model and method, including baseline approaches and our proposed \modelname. Superscripts indicate the model used: \textsuperscript{\dag} GPT-4o, \textsuperscript{\ddag} GPT-4. The baseline results are from corresponding papers. 
} 
\label{tab:model_performance}
\end{table*}

\stitle{Benchmark} 
We use the WebArena benchmark \cite{zhouwebarena}.
This benchmark comprises diverse web interaction scenarios, ranging from web shopping to customer relationship management system (CMS). The dataset consists of 812 user queries with annotated ground truth trajectories. The Webarena benchmark further provides a set of validators to programmatically validate the functional correctness of each task.

\stitle{Implementation Details} 
We choose gpt-4o\footnote{OpenAI. (2024). ChatGPT (November 20th version).} as our main LLM for both Remember and Reflect paradigm. We use the retriv\footnote{\url{https://github.com/AmenRa/retriv}} framework as the backbone of the reflective memory index, and select ``sentence-transformers/all-MiniLM-L6-v2'' as the dense embedding model for our retriever. 

\stitle{Baselines}
We compare our framework against several representative agent frameworks: (1) \textbf{ReACT} \cite{yao2023reactsynergizingreasoningacting}: a widely-used framework, which takes an observation of the environment as input, performs Chain-of-Thought reasoning, and then generates the next action. (2) \textbf{Tree-Search} \cite{koh2024tree}: an inference-time tree-search strategy to perform best-first tree-search in web environments. It enables agents to revert to the most recently validated state upon encountering a failed trajectory. (3) \textbf{LATS} \cite{zhou2024language}: a method based on Monte
Carlo tree search that
employs LLMs as agents, value functions, and optimizers for decision-making. (4) \textbf{Anticipatory Reflection} \cite{wang-etal-2024-devils}: a method that explicitly considers potential failures before action, alignment and backtracking after actions to maintain task objectives. (5) \textbf{AutoEval} \cite{panautonomous}: methods that boost agent performance using domain-general automatic evaluators. (6) \textbf{BrowserGym} \cite{dechezelles2024browsergymecosystemwebagent}: a framework that incorporates additional actions and observation tools for agents to interact with the environment.\footnote{There are other methods that use different setups, such as SteP \cite{sodhi2024step}, that employs human-engineered workflows, and AWM \cite{awm2024wang}, which uses BrowserGym framework and customizes a larger action space than standard WebArena. 
For the sake of direct comparison, these frameworks that depend on additional human efforts are not taken into comparison here.
} 


\subsection{Main Results}
\label{ssec: results}






\stitle{Overall Performance}
As shown in \Cref{tab:model_performance}, our \modelname model consistently achieves higher success rates than both Tree-Search and ReACT across all tasks. For example, on the CMS and Reddit tasks, \modelname outperforms Tree-Search by substantial margins. 
These gains demonstrate the effectiveness of combining a systematic replay buffer with a reflective memory paradigm. By leveraging past interactions, \modelname avoids repeated mistakes, leading to more accurate and efficient decisions.


\stitle{Substantial Improvements in Complex Domains}
The results in the CMS and GitLab domains are particularly notable. \modelname achieved a 30\% success rate in CMS and 28\% in GitLab, considerably higher than other tested methods. These domains often require complex navigations with web interfaces, where \modelname's capability to leverage past visited states and reflect on past actions proves especially beneficial. 

\stitle{Comparison with Reflection-based Frameworks}
When compared to complex frameworks employing sophisticated reflection mechanisms (e.g., AR, AutoEval), \modelname holds its own or exceeds performance, with a total success rate (SR) of 27.3\%. While AR and AutoEval offer robust reflection capabilities, \modelname's integrated approach to first remembering and then reflecting allows it to preemptively correct paths and further refine strategies.
The success can be attributed to the method's dual-paradigm system. We show more analysis in \Cref{ssec: ablation}.


\subsection{Ablation Study}
\label{ssec: ablation}

\begin{figure}[t]
    \centering
    \small
        \includegraphics[width=\linewidth]{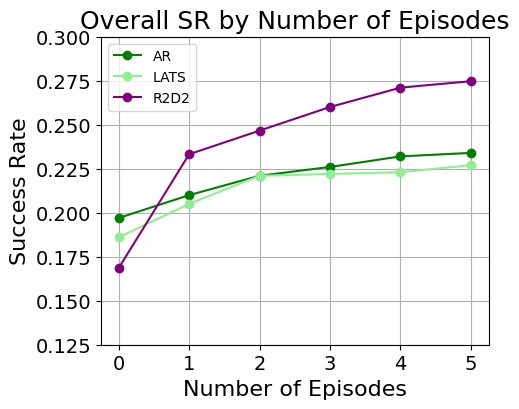}
               
        \caption{Performance comparison with different reflection-based methods. \modelname achieves marked increase at the first episode. Our manual inspection in \Cref{fig: error_analy} shows 75\% of the initial increase is attributed to fixing the navigation failures.}
        \label{fig:figure1}

\end{figure}

\stitle{Ablating rounds of execution}
To better understand the strength of our proposed framework, we compare \modelname with advanced baselines that emphasize  reflection techniques.
\Cref{fig:figure1} illustrates a marked increase in the success rate of \modelname during initial episodes. Upon manual inspection, we attribute this early performance enhancement primarily to the effective resolution of navigation failures. 

By the fifth episode, \modelname substantially outperforms AR and LATS, confirming its methodological superiority. This highlights \modelname's ability to leverage historical data and adaptive strategies effectively. While AR demonstrates commendable learning capabilities through its anticipatory reflection, it fails to match \modelname's effectiveness. LATS, in contrast, shows minimal improvement. These findings support the practical superiority of the \modelname model in dynamic learning environments.

\begin{figure}[t]
    \centering
    \small
     
    \includegraphics[width=0.85\linewidth]{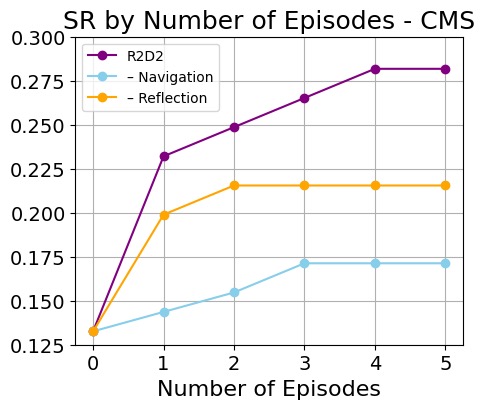}
    \caption{Performance comparison with ablation variants. Removing navigation or reflection capabilities from \modelname is very harmful to performance.}
    \label{fig:figure2}

\end{figure}
\stitle{Ablating Remember \& Reflect Paradigms} In this ablation study focused on the CMS domain, the full \modelname model substantially outperforms its variants, as shown in \Cref{fig:figure2}. The ``-- Reflection'' variant, which lacks advanced reflection capabilities, shows moderate gains, while the ``-- Navigation'' variant, which removes navigation, achieves only marginal improvement. Notably, the ``-- Reflection'' variant, though initially showing some improvement, demonstrates a limited performance increase in later episodes, suggesting that while navigation capabilities can provide early benefits, their effectiveness without reflection support plateaus quickly. This observation highlights the critical role of navigation in sustaining performance improvements over time, reinforcing that reflection alone is insufficient for long-term success in complex web environments.

\begin{table}[t]
\small
\centering
\begin{tabular}{lcc}
\toprule
\textbf{Method} & \textbf{Accuracy} & \textbf{Steps}\\
\midrule
Tree-Search~\cite{koh2024tree} & 19.2\% & 33.8 \\
AutoEval~\cite{panautonomous}    & 20.2\% & 29.2 \\
R2D2                       & \textbf{27.3\%} & \textbf{13.1} \\
\bottomrule
\end{tabular}
\caption{Comparison of task accuracy and number of actions required. R2D2 reduces the number of online steps while maintaining a higher success rate.}
\label{tab:efficiency}
\vspace{-3mm}
\end{table}

\stitle{Ablating Failed Trajectories}
To elucidate the learning dynamics of \modelname, we conduct a study to isolate the impact of failed trajectories. During this study, only successful trajectories are provided as in-context demonstrations at inference time, thereby restricting \modelname to learning exclusively from positive examples. This variant falls 6.8\% from the full implementation of \modelname to 20.5\%. This also reveals a critical limitation: the number of positive examples is insufficient to provide robust navigation and reflection to the agent during inference. Consequently, if no relevant successful trajectory is identified at retrieval time. These findings substantiate the hypothesis that failed trajectories, despite not directly addressing user queries, are instrumental in enriching \modelname's strategic repertoire, and \modelname extends beyond the mere memorization of positive examples.

\subsection{Efficiency Analysis}
Beyond improving task success rates, \modelname also demonstrates significant efficiency gains by reducing the number of online steps required per task. As shown in Table~\ref{tab:efficiency}, we compare the average steps taken to successfully address a task between \modelname and open-sourced representative baselines. \modelname completes tasks with fewer steps on average, achieving a higher success rate.\footnote{Offline memory construction for \modelname involves at most five actions per task, and the replay buffer creation is rule-based, making it lightweight.} Because web-based tasks are often bottlenecked by interactions with the live environment rather than by language model queries, minimizing the number of online steps reduces latency and overall inference time. Consequently, \modelname’s ability to leverage a cached replay buffer avoids frequent back-and-forth rollouts, leading to improved efficiency alongside its superior performance.

\subsection{Error Analysis}
\label{ssec: error_analy_4}
\begin{figure}[t]
    \centering
    \small
    \includegraphics[width=0.9\linewidth]{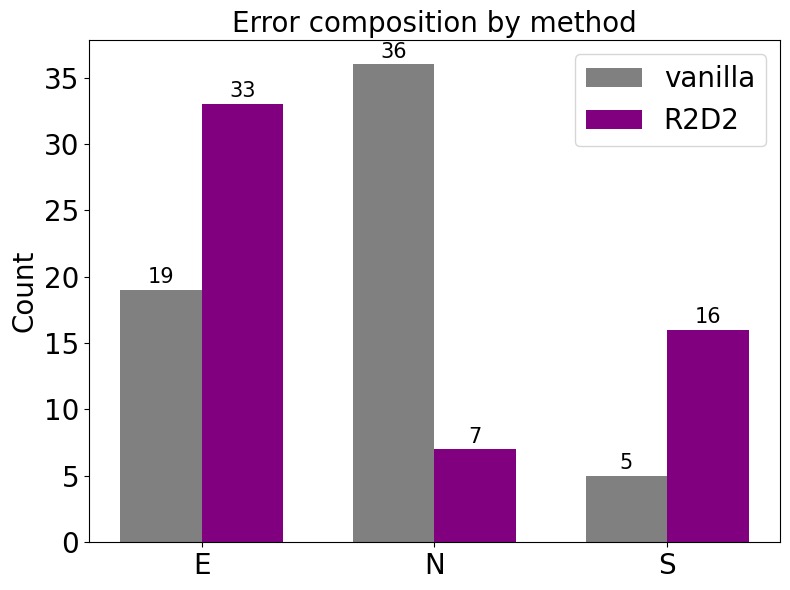}
    \vspace{-3mm}
    \caption{Error analysis of a vanilla ReACT agent's and R2D2's trajectories. ``E'' indicates execution failures, ``N'' indicates navigation failures, and ``S'' indicates success. \modelname substantially reduces navigation failures, achieving a higher success rate.}
    \label{fig: error_analy}
\end{figure}

As shown in \Cref{fig: error_analy}, we manually inspect the trajectories of the same 60 queries executed by the vanilla ReACT agent and \modelname agent.  About 60\% of the vanilla ReACT agent trajectories stall at the navigation stage, preventing the opportunity to fail in execution. In contrast, the \modelname agent substantially reduces navigation failures, reliably guiding itself toward the right content and thereby reaching a point where it is possible to fail in execution more frequently. As a result, \modelname achieves a higher pass rate overall. 
We further annotate and discuss erroneous trajectories in \Cref{sec:error_analy}. We also conduct a qualitative evaluation of the \modelname framework, as detailed in \Cref{sec: qualitative}.

\label{ssec: error_analysis}

\section{Conclusion}
The \modelname framework significantly enhances web agents' capabilities by integrating Remember and Reflect paradigms, enabling more effective navigation and interaction in complex web environments. This approach leads to measurable improvements in performance, reducing errors and increasing task completion rates. \modelname not only outperforms existing models but also offers a scalable solution adaptable to various domains. Future work could extend its application, further optimizing agent functionality across broader scenarios.

\section*{Acknowledgments}
Muhao Chen was supported by the DARPA FoundSci Grant HR00112490370, the NSF of the United States Grant ITE 2333736.  This work is partially supported by DARPA award HR00112220046. Any opinions, findings, conclusions, or recommendations expressed here are those of the authors and do not necessarily reflect the view of our sponsors.

\section*{Limitations}
\stitle{Language Studied} Our experiments were exclusively conducted in English. This limitation restricts our understanding of the model's efficacy across different linguistic contexts, potentially overlooking cultural and language-specific nuances that could affect the agent's performance in non-English web environments. 

\stitle{Focus on a Single Benchmark}
Our experiments are confined to the WebArena benchmark using GPT-4o, which may raise concerns about their broader applicability. However, WebArena spans a broad set of tasks, ranging from online shopping to social media interactions, and \modelname's strong performance across these varied scenarios suggests that our cached-search approach and Remember/Reflect paradigms are not restricted to a single domain.


\stitle{Resource Constraints}
Each complete pass through WebArena incurs a cost of approximately \$200 in GPT-4o usage, making large-scale or multi-benchmark experimentation logistically challenging. That said, our approach does not inherently depend on GPT-4o. We anticipate that future research can replicate these methods using different LLM backends or other text-based environments, suggesting that our approach is not fundamentally limited in scope or generalizability.

\bibliography{anthology,custom}

\appendix


\section{Error Analysis}
\label{sec:error_analy}
Among all the execution failures of \modelname, the errors can be classified as following:

\stitle{Pessimistic Reflection (30.3\%)} When the agent makes a mistake and enters the reflection phase, it occasionally produces overly pessimistic rationales. Instead of proposing a plausible alternative action or a corrective step—such as trying a different button or re-verifying information on the same page—the agent may hastily conclude that the service is unavailable, broken, or that no solution exists. This pessimism not only mischaracterizes the underlying issue but also inhibits effective learning from the mistake. By prematurely giving up, the agent misses opportunities to refine its approach, explore subtle variations in the action sequence, or simply retry a failed step with slight modifications.

\stitle{Lack of GUI understanding (24.2\%)} In certain scenarios, the agent struggles to properly interpret or interact with the graphical user interface (GUI) elements of the webpage. For example, when the user's query requires submitting information through an online form, the agent may fail to pinpoint the correct input fields or submission buttons, even after correctly navigating to the right page. As a result, it may click on the wrong element, repeatedly fail to submit required information, or get stuck trying to identify how to move forward.

\stitle{Difficulty with Executing Complex Plan (20.2\%)} After reaching the desired section of a website, the agent may still falter when asked to carry out intricate, multi-step tasks. For instance, it could be instructed to iterate through a list of items, adding each one to a cart, verifying their details, and then proceeding to a checkout process. This difficulty suggests that, although navigation is now more reliable, the agent still needs improved reasoning capabilities and better long-term action planning to handle scenarios that demand careful step-by-step execution.

\section{Prompt Details}
\begin{figure}[h]
    \centering
    \vspace{-3mm}
    \includegraphics[width=\linewidth]{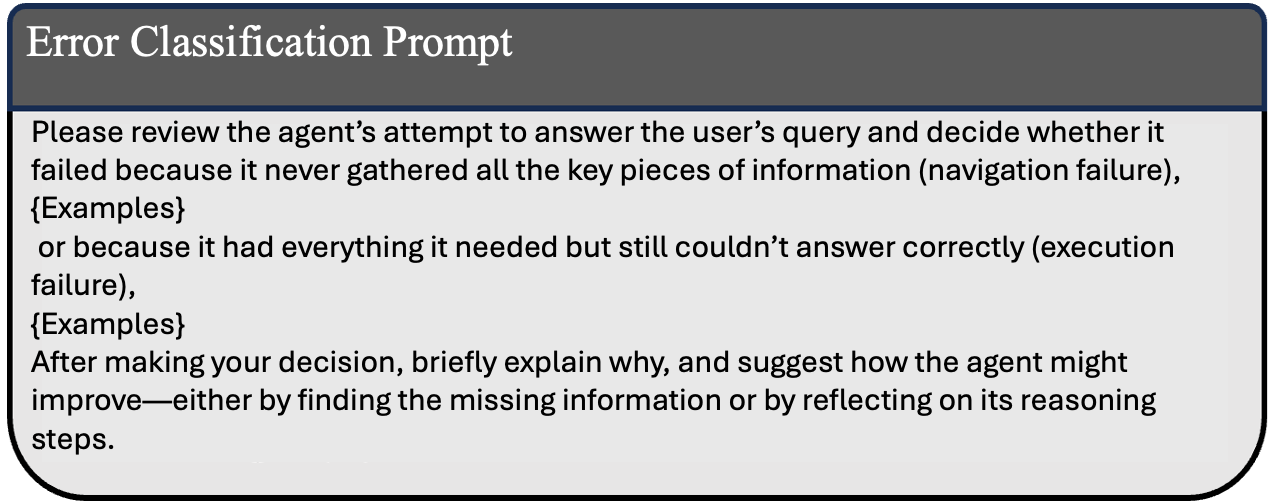}
    \vspace{-3mm}
    \caption{Error classification prompt to determine agent trajectory error type.}
    \label{fig:error_classification_p}
\end{figure}

\begin{figure}[t!]
    \centering
    \includegraphics[width=\linewidth]{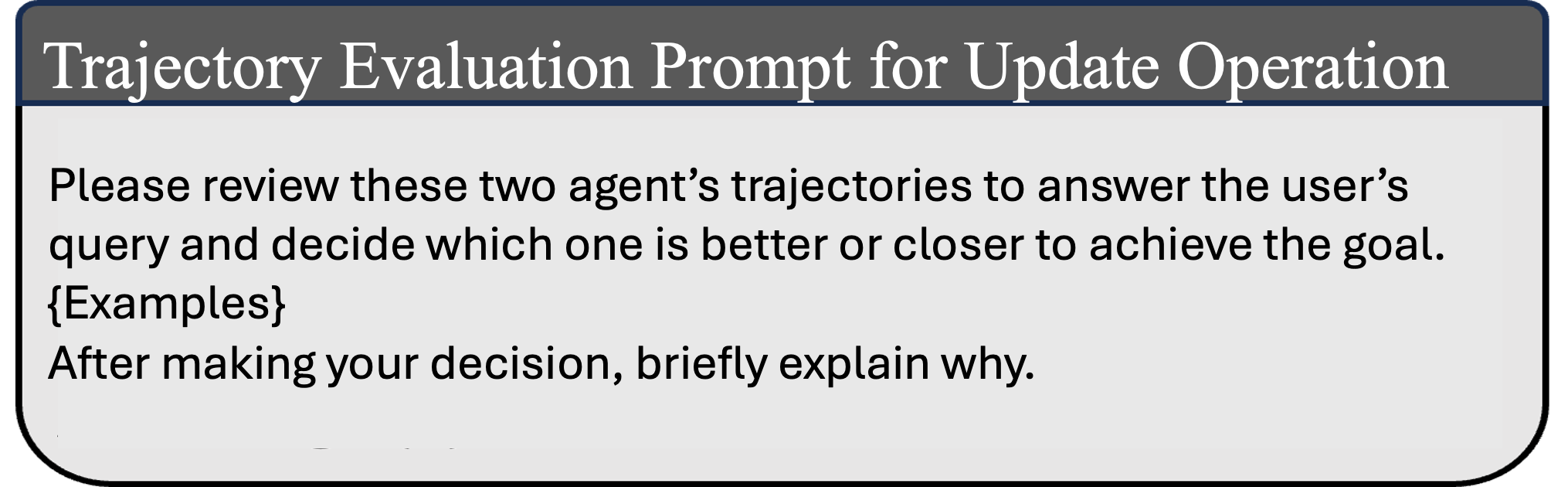}
    \vspace{-3mm}
    \caption{Trajectory evaluation prompt for Update Operation. The LLM compares two trajectory and  determines which one is better for addressing the user query.}
    \vspace{-3mm}
    \label{fig:update_eval_p}
\end{figure}

In this section, \Cref{fig:error_classification_p} and \Cref{fig:update_eval_p} show prompt details in the paradigms.

\begin{table*}[t]
\scriptsize 
\centering
\setlength{\tabcolsep}{4pt} 
\renewcommand{\arraystretch}{1.2} 

\begin{tabularx}{\textwidth}{|>{\raggedright\arraybackslash}X|>{\raggedright\arraybackslash}X|>{\raggedright\arraybackslash}X|>{\raggedright\arraybackslash}X|}
\hline
\textbf{User Query} & \textbf{gpt-4o Failed Traj.} & \textbf{R2D2 Intermediates} & \textbf{R2D2 Final Traj.} \\
\hline

\textbf{Get the billing name of the oldest complete order.} & 
1. Click 'SALES' icon  
2. Click order\_id/65  
3. Scroll down  
4. Stop action: Grace Nguyen \newline
5. \includegraphics[scale=0.025]{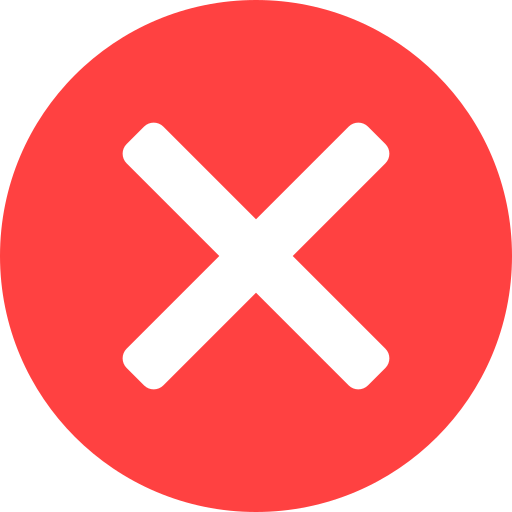}

& 
\textbf{ROUND 1} \newline
\textbf{Navigation}: 
1. Click 'SALES'  
2. Click 'Orders'  

\textbf{Reflection}: Agent fails to reach the correct page.
& 
1. Click 'SALES' icon  
2. Click 'Orders'  
3. Click '↑ Purchase Date'  
4. Click 'Filters' icon  
5. Click 'Status'  
6. Type 'Complete'  
7. Click 'Apply Filters'  
8. Click 'View'  
9. Stop action: John Lee  \newline
10. \includegraphics[scale=0.025]{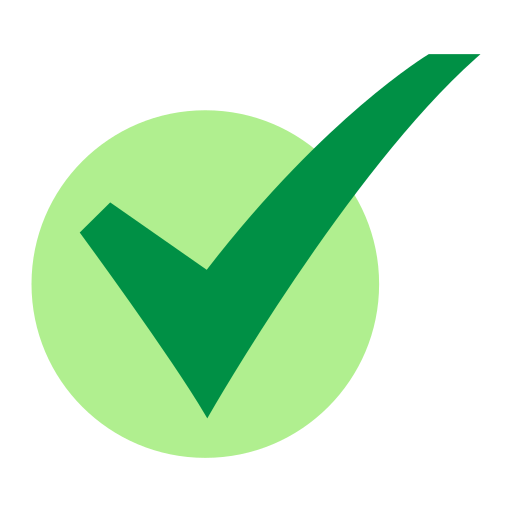}
\\
\hline

\textbf{List the top 2 search terms in my store.} & 
1. Click 'CATALOG'  
2. Click 'SALES'  
3. Click 'REPORTS'  
4. Click 'CATALOG' \newline
5. \includegraphics[scale=0.02]{figures/cross_mark.jpg}
&  
\textbf{ROUND 1} \newline
\textbf{Navigation}: 
1. Click 'REPORTS'  
2. Click 'Search Terms'  

\textbf{Reflection}: Agent fails to reach the correct page. \newline
\textbf{...} \newline
\textbf{ROUND 3} \newline
\textbf{Navigation}: Correct navigation! \newline
\textbf{Reflection}: Arrives correctly but needs to sort entries.
& 
1. Click 'REPORTS'  
2. Click 'Search Terms'  
3. Click 'Hits'  
4. Click '↓ Hits'  
5. Click '↑ Hits'  
6. Stop action: Top 2 search terms are 'hollister' (19 hits) and 'Joust Bag' (10 hits).\newline
7. \includegraphics[scale=0.02]{figures/correct_mark.jpg}
\\
\hline


\end{tabularx}

\caption{
Comparison of action trajectories for resolving user queries. This table illustrates how the GPT-4o model fails to navigate complex web interfaces, whereas the proposed \modelname framework successfully manages both intermediate steps and final actions to accurately address the same queries. 
}
\label{tab: case_study_table}
\end{table*}


\section{Qualitative Analysis}
\label{sec: qualitative}
 As shown in \Cref{tab: case_study_table}, an illustrative case can be observed in the user query asking for the billing name of the oldest complete order. The trajectory implemented by gpt-4o failed to resolve the query as it did not properly navigate the complex web interface. In contrast, the trajectory of \modelname shows a series of steps that meticulously navigate through the interface. This targeted navigation led directly to the successful completion of the task, emphasizing the necessity of precise and thoughtful navigation strategies to effectively interact with and extract information from sophisticated web environments.

In instances where navigation is executed correctly but is insufficient to solve the task, the reflection module of \modelname plays a crucial role. A clear example of this is the task to list the top two search terms in the store. While the gpt-4o trajectory navigates correctly to the 'Search Terms' section, it does not delve deeper into analyzing or sorting the data, resulting in incomplete and inaccurate information retrieval. Conversely, \modelname not only accesses the correct section but also actively manipulates the data display by sorting the search terms according to their hits, thereby precisely identifying and articulating the top search terms. This demonstrates the power of \modelname's reflective capabilities.

\section{Further Efficiency Analysis of \modelname}
The R2D2 framework's efficiency and decision-making capabilities are significantly shaped by its two core components: the ``Remember'' and ``Reflect'' paradigms. This analysis will primarily focus on the ``Remember'' paradigm, which is responsible for dynamically reconstructing the web environment, effectively building a ``map'' that the agent uses for navigation. Its complexity is characterized by the replay buffer and the $A^*$ search algorithm.

\stitle{Space Complexity of Replay Buffer} The replay buffer stores a graph where nodes represent observed web page states and edges represent transitions (actions). We would like to highlight that the replay buffer’s space complexity is $O(1)$, contending that its overall size is bounded by the number of website pages in a given domain. In practice, to ensure scalability even in potentially larger or more dynamic environments, we employ an eviction policy (e.g., removing least recently or least frequently visited nodes). This maintains a buffer of a fixed maximum size.

\stitle{Time Complexity of $A*$ Search} The $A*$ search operates on the replay buffer graph. Its time complexity is generally $O(N \log N)$ when using an efficient priority queue, where $N$ is the number of nodes (states) expanded from the replay buffer graph during a specific search process. The search process, which relies on offline memory, requires fewer steps than other online search or reflection baselines as shown in \Cref{tab:efficiency}. Because the primary time overhead is attributed to stepping within the web environment, we approximate the inference-time latency by counting the number of steps; And calls to the LLM contribute minimally to the overall time cost in our research implementation. Furthermore, in a production environment the LLM latency can be made nearly arbitrarily small, making the number of webpage steps even more significant to algorithm latency.


\end{document}